\documentclass[10pt,twocolumn,letterpaper]{article}

\usepackage{capt-of}
\usepackage{iccv}
\usepackage{times}
\usepackage{graphicx}
\usepackage{amsmath}
\usepackage{amssymb}
\usepackage{booktabs}
\usepackage{color}
\usepackage{lipsum}
\usepackage{multicol}
\usepackage{xspace}
\usepackage[pagebackref=true,breaklinks=true,letterpaper=true,colorlinks,bookmarks=false]{hyperref}
\usepackage{cleveref}
\usepackage{soul}
\usepackage{subcaption}

\iccvfinalcopy
\ificcvfinal\fi
\def\iccvPaperID{4870}

\newcommand{\bx}{\mathbf{x}}

\newcommand{\Phihat}{{\widehat\Phi}}

\newcommand*\samethanks[1][\value{footnote}]{\footnotemark[#1]}

\makeatletter
\renewcommand{\paragraph}{%
  \@startsection{paragraph}{4}%
  {\z@}{0.5em}{-1em}%
  {\normalfont\normalsize\bfseries}%
}
\makeatother

\makeatletter
\DeclareRobustCommand\onedot{\futurelet\@let@token\@onedot}
\def\@onedot{\ifx\@let@token.\else.\null\fi\xspace}
\def\eg{\emph{e.g}\onedot}

\def\etal{\emph{et al}\onedot}
\makeatother

\title{Unsupervised Learning of Landmarks by Descriptor Vector Exchange}
\author{%
James Thewlis\thanks{Equal Contribution. James was with the VGG during part of this work.}\\
\small Unitary\\
{\tt\footnotesize james@unitary.ai}
\and
Samuel Albanie\samethanks\\
\small VGG, University of Oxford\\
{\tt\footnotesize albanie@robots.ox.ac.uk}
\and
Hakan Bilen\\
\small University of Edinburgh\\
{\tt\footnotesize hbilen@ed.ac.uk}
\and
Andrea Vedaldi\\
\small VGG, University of Oxford\\
{\tt\footnotesize vedaldi@robots.ox.ac.uk}
}

\begin{document}
\maketitle

\begin{abstract}
Equivariance to random image transformations is an effective method to learn landmarks of object categories, such as the eyes and the nose in faces, without manual supervision.
However, this method does not explicitly guarantee that the learned landmarks are consistent with changes between different instances of the same object, such as different facial identities.
In this paper, we develop a new perspective on the equivariance approach by noting that dense landmark detectors can be interpreted as local image descriptors equipped with invariance to intra-category variations.
We then propose a direct method to enforce such an invariance in the standard equivariant loss.
We do so by exchanging descriptor vectors between images of different object instances prior to matching them geometrically.
In this manner, the same vectors must work regardless of the specific object identity considered.
We use this approach to learn vectors that can simultaneously be interpreted as local descriptors and dense landmarks, combining the advantages of both.
Experiments on standard benchmarks show that this approach can match, and in some cases surpass state-of-the-art performance
amongst existing methods that learn landmarks without supervision. Code is available at \mbox{\url{www.robots.ox.ac.uk/~vgg/research/DVE/}}.
\end{abstract}

\section{Introduction}\label{s:intro}

Learning without manual supervision remains an open problem in machine learning and computer vision.
Even recent advances in self-supervision~\cite{jenni2018self,kanezaki2018rotationnet}
are often limited to learning generic feature extractors and still require some manually annotated data to solve a concrete task such as landmark detection.
In this paper, we thus consider the problem of learning the landmarks of an object category, such as the eyes and nose in faces, without any manual annotation.
Namely, given as input a collection of images of a certain object, such as images of faces, the goal is to learn what landmarks exist and how to detect them.

In the absence of manual annotations, an alternative supervisory signal is required.
Recently,~\cite{thewlis17unsupervised} proposed to build on the fact that landmark detectors are \emph{equivariant} to image transformations.
For example, if one translates or rotates a face, then the locations of the eyes and nose follow suit.
Equivariance can be used as a learning signal by applying random synthetic warps to images of an object and then requiring the landmark detector to be consistent with these transformations.

The main weakness of this approach is that equivariance can only be imposed for transformations of \emph{specific} images.
This means that a landmark detector can be perfectly consistent with transformation applied to a specific face and still match an eye in a person and the nose in another.
In this approach, achieving consistency across object instances is left to the generalisation capabilities of the underlying learning algorithm.

\begin{figure}[t]
\centering\includegraphics[trim={0.5cm 0 0 0},clip,width=0.9\columnwidth]{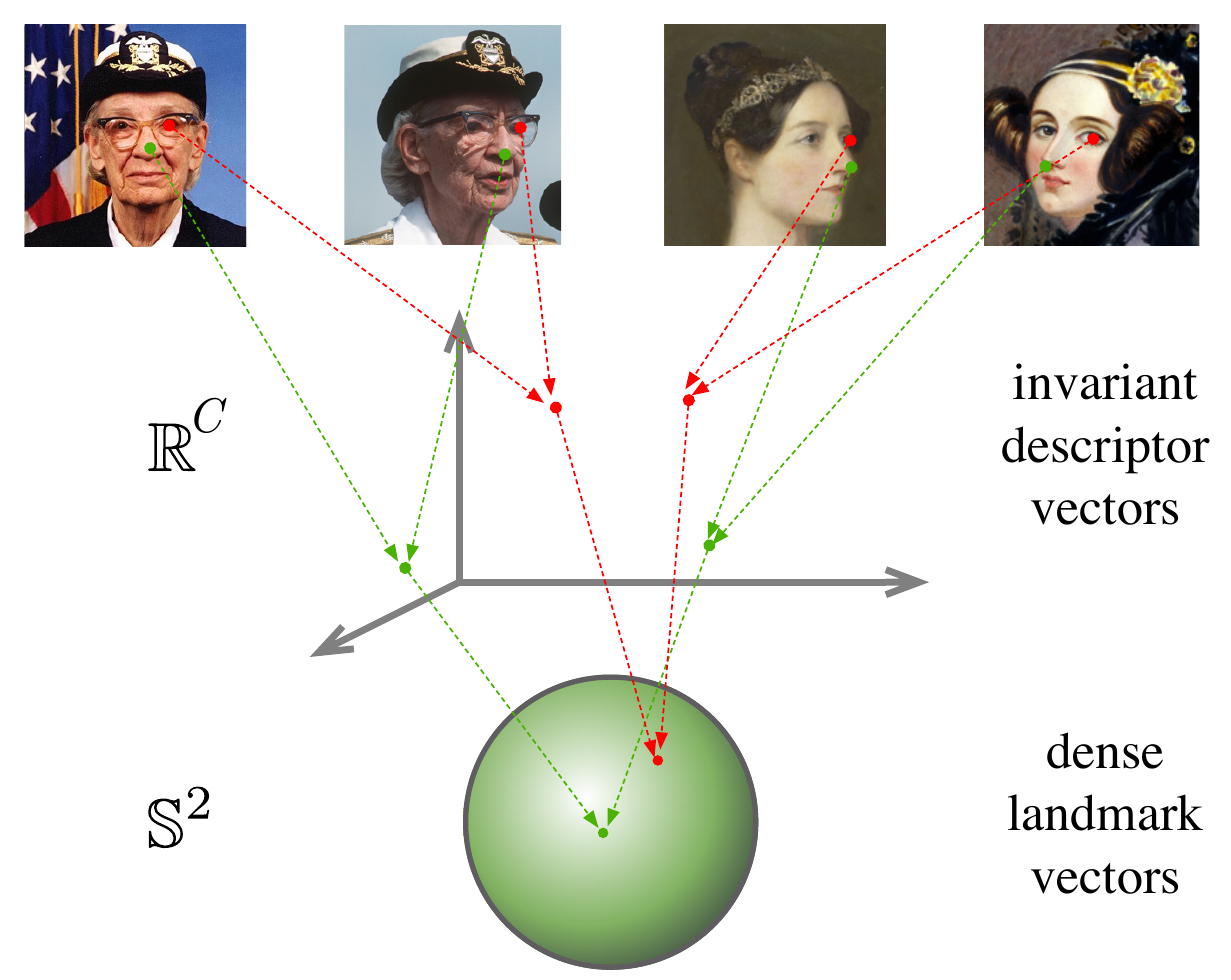}
\caption{\textbf{Descriptor-landmark hierarchy.}
A local invariant descriptor maps image pixels to distinctive vectors that are invariant to viewing conditions such as a viewpoint.
A dense landmark detector maps pixels to unique points of the object's surface, such as eyes and nose in faces, to points on the surface of a sphere.
Both produce invariant and distinctive vectors, but landmarks are also invariant to intra-class variations.
Taken together, they represent a hierarchy of distinctive pixel embeddings of increasing invariance.
}\label{f:hier}
\vspace{-0.3cm}
\end{figure}

In this paper, we offer a new perspective on the problem of learning landmarks, generalising prior work and addressing its shortcomings.
We start by establishing a link between two apparently distinct concepts: landmarks and local image descriptors (\cref{f:hier}).
Recall that a descriptor, such as SIFT, is a vector describing the appearance of the image around a given point.
Descriptors can establish correspondences between images because they are invariant to viewing effects such as viewpoint changes.
However, similar to descriptors, landmarks can also establish image correspondences by matching concepts such as eyes or noses detected in different images.

Thus invariant descriptors and landmark detectors are similar, but landmarks are invariant to intra-class variations in addition to viewing effects.
We can make this analogy precise if we consider \emph{dense} descriptors and landmarks~\cite{thewlis17Bunsupervised,alp2017densereg,shu2018deforming}.
A dense descriptor associates to each image pixel a $C$-dimensional vector, whereas a dense landmark detector associates to each pixel a 2D vector, which is the index of the landmark in a $(u,v)$ parameterisation of the object surface.
Thus we can interpret a landmark as a tiny 2D descriptor.
Due to its small dimensionality, a landmark loses the ability to encode instance-specific details of the appearance, but gains robustness to intra-class variations.

Generalising this idea, we note that any invariant descriptor can be turned into a landmark detector by equipping it with robustness to intra-class variations.
Here we propose a new method that can do so \emph{without} reducing the dimensionality of the descriptor vectors.
The formulation still considers pairs of synthetically-transformed images as~\cite{thewlis17Bunsupervised} do, but this time landmarks are represented by arbitrary $C$-dimensional vectors.
Then, before geometric consistency (equivariance) is enforced, the landmark vectors extracted from one image \emph{are exchanged with similar vectors extracted from other random images of the object}.
This way geometric consistency between an image and its transformations can only be achieved if vectors have an intra-class validity, and thus effectively characterise landmarks.

Empirically (\cref{s:experiments}), we show that the key advantage of this formulation, which we term Descriptor Vector Exchange (DVE), is that it produces embedding vectors that simultaneously work well as instance-specific image descriptors \emph{and} landmarks, capturing in a single representation the advantages of both, and validating our intuition.

\section{Related work}\label{s:related}

\noindent \textbf{General image matching.} Image matching based on local features has been an extensively studied problem in the literature with applications to wide-baseline stereo matching~\cite{pritchett1998wide} and image retrieval~\cite{tuytelaars1999matching}.
The generic pipeline contains the following steps:
i) detecting a sparse set of interest points~\cite{mikolajczyk2005comparison} that are covariant with a class of transformations,
ii) extracting local descriptors (\eg~\cite{lowe2004distinctive,tola2008fast}) at these points that are invariant to viewpoint and illumination changes, and
iii) matching the nearest neighbour descriptors across images with an optional geometric verification.
While the majority of the image matching methods rely on hand-crafted detectors and descriptors, recent work show that CNN-based models can successfully be trained to detect covariant detectors~\cite{lenc16learning} and invariant descriptors~\cite{zagoruyko2015learning,paulin2015local}.
We build our method on similar principles, covariance and invariance, but with an important difference that it can learn intrinsic features for object categories in contrast to generic ones.

\noindent \textbf{Cross-instance object matching.} The SIFT Flow method~\cite{Liu2011} extends the problem of finding dense correspondences between same object instances to different instances by matching their SIFT features~\cite{lowe2004distinctive} in a variational framework.
This work is further improved by using multi-scale patches~\cite{hassner2012sifts}, establishing region correspondences~\cite{ham2016proposal} and replacing SIFT features with CNN ones~\cite{long2014convnets}.
In addition, Learned-Miller~\cite{learned2006data} generalises the dense correspondences between image pairs to an arbitrary number of images by continuously warping each image via a parametric transformation.
RSA~\cite{peng2012rasl}, Collection Flow~\cite{Kemelmacher-Shlizerman2012} and Mobahi~\etal~\cite{Mobahi2014} project a collection of images into a lower dimensional subspace and perform a joint alignment among the projected images.
AnchorNet~\cite{novotny17learning} learns semantically meaningful parts across categories, although is trained with image labels.

\noindent \textbf{Transitivity.} The use of transitivity to regularise structured data has been proposed by several authors~\cite{sundaram2010dense,zach2010disambiguating,zhou15flowweb,zhou16Blearning} in the literature.
Earlier examples~\cite{sundaram2010dense,zach2010disambiguating} employ this principle to achieve forward-backward consistency in object tracking and to identify inconsistent geometric relations in structure from motion respectively.
Zhou~\etal~\cite{zhou15flowweb,zhou16Blearning} enforce a geometric consistency to jointly align image sets or supervise deep neural networks in dense semantic alignment by establishing a cycle between each image pair and a 3D CAD model.
DVE also builds on the same general principle of transitivity, however, it operates in the space of appearance embeddings in contrast to verification of subsequent image warps to a composition.

\noindent \textbf{Unsupervised learning of object structure.} Visual object characterisation (\eg~\cite{Cootes1995,Fergus2003,leibe2004combined,Dalal2005,Felzenszwalb2010}) has a long history in computer vision with extensive work in facial landmark detection and human body pose estimation.
A recent unsupervised method that can learn geometric transformations to optimise classification accuracy is the spatial transformer network~\cite{Jaderberg2015}.
However, this method does not learn any explicit object geometry.
Similarly, WarpNet~\cite{kanazawa16warpnet} and geometric matching networks~\cite{rocco17} train neural networks to predict relative transformations between image pairs.
These methods are limited to perform only on image pairs and do not learn an invariant geometric embedding for the object.
Most related to our work,~\cite{thewlis17unsupervised} characterises objects by learning landmarks that are consistent with geometric transformations without any manual supervision, while~\cite{novotny2018self} similarly use such transformations for semantic matching. 
The authors of~\cite{thewlis17unsupervised} extended their approach to extract a dense set of landmarks by projecting the raw pixels on a surface of a sphere in~\cite{thewlis17Bunsupervised}.
Similar work~\cite{schmidt2017self} leverages frame-to-frame correspondence using Dynamic Fusion~\cite{newcombe2015dynamicfusion} as supervision to learn a dense labelling for human images.
We build our method, DVE, on these approaches and further extend them in significant ways.
First, we learn more versatile descriptors that can encode both generic and object-specific landmarks and show that we can gradually learn to move from generic to specific ones.
Second, we improve the cross-instance generalisation ability by better regularising the embedding space with the use of transitivity.
Finally, we show that DVE both qualitatively and quantitatively outperforms~\cite{thewlis17unsupervised,thewlis17Bunsupervised} in facial landmark detection (\cref{s:experiments}).
Recent work~\cite{zhang2018unsupervised,jakab2018unsupervised,Wiles18a,shu2018deforming} proposes to disentangle appearance from pose by estimating dense deformation field~\cite{Wiles18a,shu2018deforming} and by learning landmark positions to reconstruct one sample from another.
We compare DVE to these approaches in \cref{s:experiments}.

\section{Method}\label{s:method}

\begin{figure*}[t]
\vspace*{-0.7cm}\hspace*{0.4cm}\centering\includegraphics[width=1\textwidth,keepaspectratio]{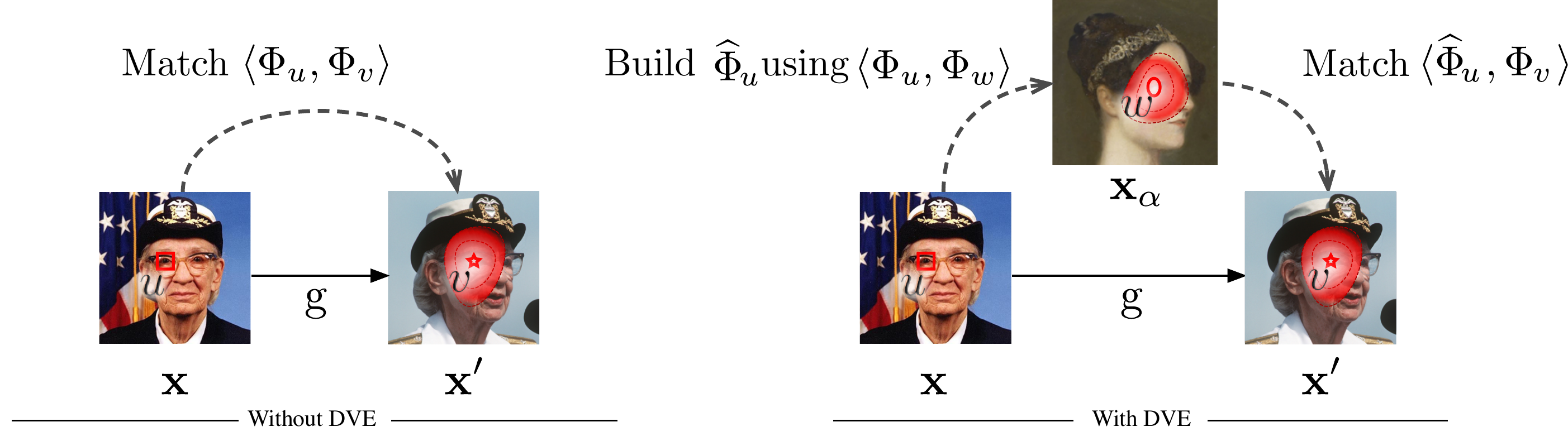}
\caption{%
We learn a dense embedding $\Phi_u(\bx)\in\mathbb{R}^C$ of image pixels.
The embedding is learned from pairs of images $(\bx,\bx')$ related by a known warp $v=g(u)$. Note that in practice, we do not have access to pairs of pairs of images with a known correspondence---thus, throughout this work the warps are generated synthetically.
Left: the approach of~\cite{thewlis17Bunsupervised} directly matches embedding $\Phi_u(\bx)$ from the left image to embeddings $\Phi_v(\bx')$ in the right image.
Right: DVE replaces $\Phi_u(\bx)$ from its reconstruction $\hat \Phi_u(\bx|\bx_\alpha)$ obtained from the embeddings in a third auxiliary image $\bx_\alpha$.
Importantly, the correspondence with $\bx_\alpha$ does not need to be known. 
}\label{f:transv}
\end{figure*}

We first summarise the method of~\cite{thewlis17Bunsupervised} and then introduce DVE, our extension to their approach.

\subsection{Learning dense landmarks using equivariance}\label{s:die}

Denote by $\bx\in\mathbb{R}^{3\times H\times W}$ an image of an object, by $\Omega=\{0,\dots,H-1\}\times\{0,\dots,W-1\}$ its domain, and by $u\in \Omega$ an image pixel.
Consider as in~\cite{thewlis17Bunsupervised} a spherical parameterisation of the object surface, where each point on the sphere indexes a different characteristic point of the object, i.e.~a landmark.
Our goal is to learn a function $\Phi$ that maps pixels $u \in \Omega$ to their corresponding landmark indices $\Phi_u(\bx) \in \mathbb{S}^2$.

The authors of~\cite{thewlis17Bunsupervised} showed that $\Phi$ can be learned without manual supervision by requiring it to be \emph{invariant} with transformations of the image.
Namely, consider a random warp $g : \Omega \rightarrow \Omega$ and denote with $g\bx$ the result of applying the warp to the image.\footnote{I.e.~$(g\bx)_u = \bx_{g^{-1}u}$.}
Then, if the map assigns label $\Phi_{u}(\bx)$ to pixel $u$ of image $\bx$, it must assign the same label $\Phi_{gu}(g\bx)$ to pixel $gu$ of the deformed image $g\bx$.
This is because, by construction, pixels $u$ and $gu$ land on the same object point, and thus contain the same landmark.
Hence, we obtain the \emph{equivariance constraint} $\Phi_{u}(\bx)=\Phi_{gu}(g\bx)$.

This version of the equivariance constraint is not quite sufficient to learn meaningful landmarks.
In fact, the constraint can be satisfied trivially by mapping all pixels to some fixed point on the sphere.
Instead, we must also require landmarks to be \emph{distinctive}, i.e.~to identify a unique point in the object.
This is captured by the equation:
\begin{equation}\label{e:distinctive}
   \forall u, v\in \Omega: \quad v=gu ~\Leftrightarrow~ \Phi_{u}(\bx) = \Phi_{v}(g\bx).
\end{equation}

\paragraph{Probabilistic formulation.}

For learning, \cref{e:distinctive} is relaxed probabilistically (\cref{f:transv}). Given images $\bx$ and $\bx'$, define the probability of pixel $u$ in image $\bx$ matching pixel $v$ in image $\bx'$ by normalising the cosine similarity $\langle\Phi_u(\bx),\Phi_v(\bx')\rangle$ of the corresponding landmark vectors:
\begin{equation}\label{e:probmatch}
p(v|u;\Phi,\bx,\bx') =
\frac
{e^{\langle \Phi_u(\bx),\Phi_v(\bx') \rangle}}
{\int_\Omega e^{\langle \Phi_u(\bx),\Phi_{t}(\bx') \rangle}\,dt}.
\end{equation}
Given a warp $g$, and image $\bx$ and its deformation $\bx' = g\bx$, constraint~\cref{e:distinctive} is captured by the loss:
\begin{equation}\label{e:loss1}
\mathcal{L}(\Phi;\bx,\bx',g)
=
\frac{1}{|\Omega|^{2}}
\int_\Omega \int_\Omega \|v - gu\|\, p(v|u;\Phi,\bx,\bx') \,du\,dv
\end{equation}
where $\|v - gu\|$ is a distance between pixels.
In order to understand this loss, note that $\mathcal{L}(\Phi;\bx,\bx',g)=0$ if, and only if, for each pixel $u\in\Omega$, the probability $p(v|u;\Phi,\bx,\bx')$ puts all its mass on the corresponding pixel $gu$.
Thus minimising this loss encourages $p(v|u;\Phi,\bx,\bx')$ to establish correct deterministic correspondences.

Note that the spread of probability~\eqref{e:probmatch} only depends on the angle between landmark vectors.
In order to allow the model to modulate this spread directly, the range of function $\Phi$ is relaxed to be $\mathbb{R}^3$.
In this manner, estimating longer landmark vectors causes~\eqref{e:probmatch} to become more concentrated, and this allows the model to express the confidence of detecting a particular landmark at a certain image location.\footnote{The landmark identity is recovered by normalising the vectors to unit length.}

\paragraph{Siamese learning with random warps.}

We now explain how~\eqref{e:loss1} can be used to learn the landmark detector function $\Phi$ given only an unlabelled collection $\mathcal{X}=\{\bx_1,\dots,\bx_n\}$ of images of the object.
The idea is to synthesise for each image a corresponding random warp from a distribution $\mathcal{G}$.
Denote with $\mathcal{P}$ the empirical distribution over the training images; then this amounts to optimising the energy
\begin{equation}\label{e:energy}
E(\Phi)
=
\mathbb{E}_{\bx \sim \mathcal{P}, g \sim \mathcal{G}}
\left[
\mathcal{L}(\Phi;\bx,g\bx,g)
\right].
\end{equation}
Implemented as a neural network, this is a Siamese learning formulation because the network $\Phi$ is evaluated on both $\bx$ and $g\bx$.

\subsection{From landmarks to descriptors}\label{s:capacity}

\Cref{e:distinctive} says that landmark vectors must be invariant to image transformations and distinctive.
Remarkably, exactly the same criterion is often used to define and learn local invariant feature descriptors instead~\cite{brown10discriminative}.
In fact, if we relax the function $\Phi$ to produce embeddings in some high-dimensional vector space 
$\mathbb{R}^C$, then the formulation above can be used out-of-the-box to learn descriptors instead of landmarks.

Thus the only difference is that landmarks are constrained to be tiny vectors (just points on the sphere), whereas descriptors are usually much higher-dimensional.
As argued in~\cref{s:intro}, the low dimensionality of the landmark vectors forgets instance-specific details and promotes intra-class generalisation of these descriptors.

The opposite is also true: we can start from any descriptor and turn it into a landmark detector by promoting intra-class generalisation.
Using a low-dimensional embedding space is a way to do so, but not the only one, nor the most direct.
We propose in the next section an alternative approach.

\subsection{Vector exchangeability}\label{s:exchange}

We now propose our method, Descriptor Vector Exchange, to learn embedding vectors that are distinctive, transformation invariant, and insensitive to intra-class variations, and thus identify object landmarks.
The idea is to encourage the \emph{sets} of embedding vectors extracted from an image to be exchangeable with the ones extracted from another while retaining matching accuracy.

In more detail, let $(\bx,\bx',g)$ be a warped image pair (hence $\bx'=g\bx$).
Furthermore, let $\bx_\alpha$ be an \emph{auxiliary image}, containing an object of the same category as the pair $(\bx,\bx')$, but possibly a different instance.
If the embedding function $\Phi_u(\bx)$ is insensitive to intra-class variations, then the set of embedding vectors $\{\Phi_u(\bx):u\in\Omega\}$ and $\{\Phi_u(\bx_\alpha):u\in\Omega\}$ extracted from any two images should be approximately the same.
This means that, in loss~\eqref{e:loss1}, we can exchange the vectors $\Phi_u(\bx)$ extracted from image $\bx$ with corresponding vectors extracted from the auxiliary image $\bx_\alpha$.

Next, we integrate this idea in the probabilistic learning formulation given above (\cref{f:transv}).
We start by matching pixels in the source image $\bx$ to pixels in the auxiliary image $\bx_\alpha$ by using the probability $p(w|u;\Phi,\bx,\bx_\alpha)$ computed according to~\cref{e:probmatch}.
Then, we reconstruct the source embedding $\Phi_u(\bx)$ as the weighted average of the embeddings $\Phi_{w}(\bx_\alpha)$ in the auxiliary image, as follows:
\begin{equation}\label{e:rec}
   \Phihat_u(\bx|\bx_{\alpha})
   =
   \int \Phi_{w}(\bx_\alpha)p(w|u;\Phi,\bx,\bx_\alpha)\,dw.
\end{equation}
Once $\Phihat_u$ is computed, we use it to establish correspondences between $\bx$ and $\bx'$, using~\cref{e:probmatch}.
This results in the matching probability:
\begin{equation}\label{e:probmatch2}
p(v|u;\Phi,\bx,\bx',\bx_\alpha) =
\frac
{e^{\langle\Phihat_u(\bx|\bx_\alpha),\Phi_v(\bx') \rangle}}
{\int_\Omega e^{\langle \Phihat_u(\bx|\bx_\alpha),\Phi_{t}(\bx') \rangle}\,dt}.
\end{equation}
This matching probability can be used in the same loss function~\eqref{e:loss1} as before, with the only difference that now each sample depends on $\bx,\bx'$ as well as the auxiliary image $\bx_\alpha$.

\paragraph{Discussion.}

While this may seem a round-about way of learning correspondences, it has two key benefits: as~\cref{e:loss1} encourages vectors to be invariant and distinctive; in addition to~\cref{e:loss1}, DVE also requires vectors to be compatible between different object instances.
In fact, without such a compatibility, the reconstruction~\eqref{e:rec} would result in a distorted, unmatchable embedding vector.
Note that the original formulation of~\cite{thewlis17Bunsupervised} lacks the ability to enforce this compatibility directly.

\subsection{Using multiple auxiliary images}\label{s:group}

A potential issue with~\cref{e:probmatch2} is that, while image $\bx'$ can be obtained from $\bx$ by a synthetic warp so that all pixels can be matched, image $\bx_\alpha$ is only weakly related to the two.
For example, partial occlusions or out of plane rotations may cause some of the pixels in $\bx$ to not have corresponding pixels in $\bx_\alpha$.

In order to overcome this issue, we take inspiration from the recent method of~\cite{zhuang17attend} and consider not one, but a small set $\{\bx_\alpha : \alpha \in A\}$ of auxiliary images.
Then, the summation in~\cref{e:rec} is extended not just over spatial locations, but also over images in this set.
The intuition for this approach is that as long as at least one image in the auxiliary image set matches $\bx$ sufficiently well, then the reconstruction will be reliable.

\section{Experiments}\label{s:experiments}

Using datasets of human faces (\cref{s:human}), animal faces (\cref{s:cross}) and a toy robotic arm (\cref{s:robo}), we demonstrate the effectiveness of the proposed Descriptor Vector Exchange technique in two ways.
First, we show that the learned embeddings work well as visual descriptors, matching reliably different views of an object instance.
Second, we show that they \emph{also} identify a dense family of object landmarks, valid not for one, but for all object instances in the same category.
Note that, while the first property is in common with traditional and learned descriptors in the spirit of SIFT, the second clearly sets DVE embeddings apart from these.

\paragraph{Implementation details.}\label{s:details}

In order to allow for a comparison with the literature, we perform experiments with the deep neural network architecture of~\cite{thewlis17Bunsupervised} (which we refer to as \textit{SmallNet}).  Inspired by the success of the \textit{Hourglass} model in  \cite{zhang2018unsupervised}, we also experiment with a more powerful hourglass design (we use the ``Stacked Hourglass'' design of \cite{newell2016stacked} with a single stack).  The weights of both models are learned from scratch using the Adam optimiser~\cite{kingma2014adam} for 100 epochs with an initial learning rate of 0.001 and without weight decay.  Further details of the architectures are provided in the supplementary material.

\subsection{Human faces}\label{s:human}

First, we consider two standard benchmark datasets of human faces: CelebA~\cite{liu2015faceattributes} and MAFL~\cite{Zhang2016}, which is a subset of the former.
The CelebA~\cite{liu2015faceattributes} dataset contains over 200k faces of celebrities; we use the former for training and evaluate embedding quality on the smaller MAFL~\cite{Zhang2016} (19,000 train images, 1,000 test images). Annotations are provided for the eyes, nose and mouth corners.
For training, we follow the same procedure used by~\cite{thewlis17Bunsupervised} and exclude any image in the CelebA training set that is also contained in the MAFL test set.
Note that we use MAFL annotations only for evaluation and never for training of the embedding function.

We use formulation~\eqref{e:probmatch2} to learn a dense embedding function $\Phi$ mapping an image $\bx$ to $C$-dimensional pixel embeddings, as explained above.
Note that loss~\eqref{e:loss1} requires sampling transformations $g\in\mathcal{G}$; in order to allow a direct comparison with~\cite{thewlis17Bunsupervised}, we use the same random Thin Plate Spline (TPS) warps as they use, obtaining warped pairs $(\bx,\bx'=g\bx)$.  We also sample at random one or more auxiliary images $\bx_{\alpha}$ from the training set in order to implement {DVE}.

We consider several cases; in the first, we set $C=3$ and sample no auxiliary images, using formulation~\eqref{e:probmatch}, which is the same as~\cite{thewlis17Bunsupervised}.
In the second case, we set $C=16,32,64\gg 3$ but still do not use {DVE}; in the last case, we use $C=3,16,32,64$ and also use DVE.

\paragraph{Qualitative results.}

\begin{figure}[h]
\centering\includegraphics[width=0.9\columnwidth]{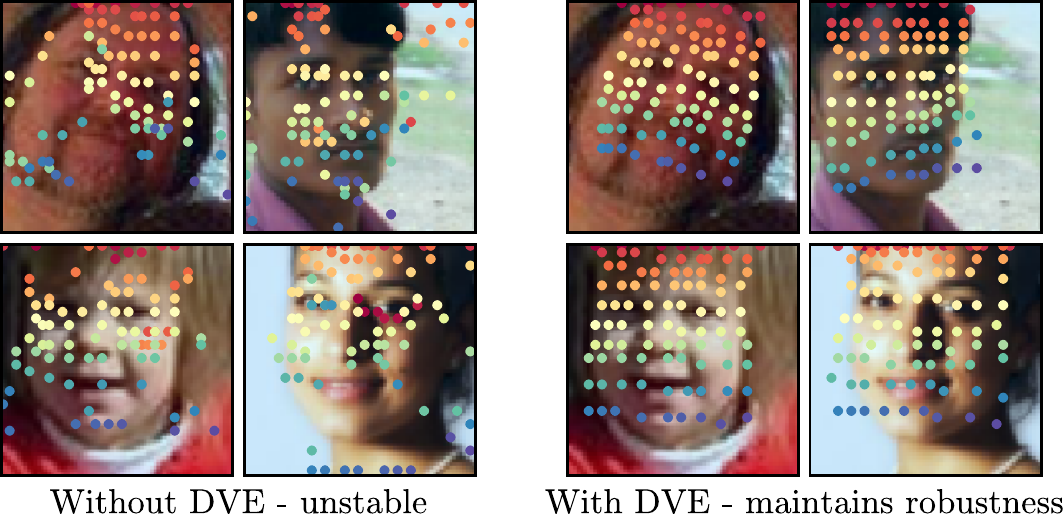}
\caption{Learning 64D descriptors without/with DVE}\label{f:qualres}
\vspace{-0.48cm}
\end{figure}

In~\cref{f:qualres} we compute 64D embeddings with SmallNet models trained with or without DVE on  AFLW$_M$ images, visualising as in ~\cref{f:splash} (left).  With DVE, matches are accurate despite large intra-class variations.  Without DVE, embedding quality degrades significantly. This shows that, by having a category-wide validity, embeddings learned with DVE identify object landmarks rather than mere visual descriptors of local appearance.

\begin{table}[t]
\setlength\tabcolsep{9pt}
\centering
\begin{tabular}{ccccc}
\toprule
Embedding & \multicolumn{2}{c}{Same identity} & \multicolumn{2}{c}{Different identity} \\
dimension & \cite{thewlis17Bunsupervised} & + DVE & \cite{thewlis17Bunsupervised} & + DVE \\
\midrule

3  & 1.33 & 1.36 & 2.89 & 3.03  \\
16 & 1.25 & 1.28 & 5.65 & 2.79  \\
32 & 1.26 & 1.29 & 5.81 & 2.79  \\
64 & 1.25 & 1.28 & 5.68 & 2.77  \\
\bottomrule
\end{tabular}

\medskip
\vspace{-0.4em}
\caption{Pixel error when matching annotated landmarks across 1000 pairs of images from CelebA (MAFL test set).}
\vspace{-0.4em}

\label{t:facematching}
\end{table}

\paragraph{Matching results.}

Next, we explore the ability of the embeddings learned with \textit{SmallNet} to match face images.
We sample pairs of different identities using MAFL test (1000 pairs total) and consider two cases:  First, we match images $\bx,\bx'$ of the \emph{same identity}; since multiple images of the same identity are not provided, we generate them with warps as before, so that the ground-truth correspondence field $g$ is known.
We extract embeddings at the annotated keypoint positions from $\bx$ and match them to their closest neighbour embedding in image $\bx'$ (searching all pixels in the target).
Second, we match images of \emph{different identities}, again using the annotations.
In both cases, we report the mean pixel matching error from the ground truth.

Examining the results in~\cref{t:facematching} we note several facts.
When matching the same identities, higher dimensional embeddings work better than lower (i.e. 3D), including in particular~\cite{thewlis17Bunsupervised}.
This is expected as high dimensional embeddings more easily capture instance-specific details; also as expected, DVE does not change the results much as here there are no intra-class variations.
When matching different identities, high-dimensional embeddings are rather poor: these descriptors are too sensitive to instance-specific details and cannot bridge intra-class variations correctly.
This justifies the choice of a low dimensional embedding in~\cite{thewlis17Bunsupervised} as the latter clearly generalises better across instances.
However, once DVE is applied, the performance of the high-dimensional embeddings is much improved, and is in fact better than the low-dimensional descriptors even for intra-class matching~\cite{thewlis17Bunsupervised}.

Overall, the embeddings learned with DVE have both better intra-class and intra-instance matching performance than~\cite{thewlis17Bunsupervised}, validating our hypothesis and demonstrating that our method for regularising the embedding is preferable to simply constraining the embedding dimensionality.

\paragraph{Landmark regression.}\label{s:landmark}

\begin{table}[t]
\setlength\tabcolsep{1pt}
\centering

\begin{tabular}{l@{\hskip -0.3cm}c@{\hskip -0.1cm}cccc}
\toprule
Method                      & Unsup.\ \,      & \,MAFL    & \,AFLW$_M$ & \,AFLW$_R$     & \,\,300W  \\
\midrule
TCDCN~\cite{Zhang2016}      &$\times$      & 7.95    &  7.65  & --        & 5.54  \\
RAR~\cite{Xiao2016}         &$\times$      &         & 7.23 & -- & 4.94  \\
MTCNN~\cite{zhang2014facial,zhang2018unsupervised}  &$\times$      &   5.39  & 6.90  & --  &  -- \\
Wing Loss~\cite{feng2018wing}$\ast$    &$\times$      &   -  & -  & -  &  4.04 \\                                                                       
\midrule
Sparse~\cite{thewlis17unsupervised}  & $\checkmark$ & 6.67 &  10.53 & -- & 7.97 \\
Structural Repr.~\cite{zhang2018unsupervised} & $\checkmark$ & 3.15 & -- & 6.58 & -- \\
FAb-Net~\cite{Wiles18a}$\ddagger$ & $\checkmark$  & 3.44 & -- & -- & 5.71\\
Def.~AE~\cite{shu2018deforming} & $\checkmark$ & 5.45 & -- & --  & --\\
Cond.~ImGen.~\cite{jakab2018unsupervised} & $\checkmark$ & \textbf{2.54} & -- & \textbf{6.31}  & --\\
UDIT~\cite{jakab2019learning}$\dagger$ & $\checkmark$ & - & - & - & 5.37 \\
Dense 3D~\cite{thewlis17Bunsupervised} & $\checkmark$ & 4.02 & 10.99 & 10.14 & 8.23\\
DVE SmallNet-64D  & $\checkmark$                  & 3.42 & 8.60 & 7.79 & 5.75  \\
DVE Hourglass-64D  & $\checkmark$ & 2.86 & \textbf{7.53} & 6.54 & \textbf{4.65}  \\
\bottomrule
\end{tabular}
\medskip
\caption{Landmark detection results on the MAFL, 300W and AFLW (AFLW$_M$ and ALFW$_R$ splits---see~\cref{s:landmark} for details). The results are reported as percentage of inter-ocular distance. $\ast$ report a more conservative evaluation metric (see~\cite{feng2018wing}),  $\dagger$ and $\ddagger$ use different training data: VoxCeleb~\cite{Nagrani17} and VoxCeleb+ (the union of VoxCeleb and VoxCeleb2~\cite{Chung18b}) respectively.
\label{t:landmark}}
\end{table}

Next, as in~\cite{thewlis17Bunsupervised} and other recent papers, we assess quantitatively how well our embeddings correspond to manually-annotated landmarks in faces.
For this, we follow the approach of~\cite{thewlis17Bunsupervised} and add on top of our embedding 50 filters of dimension $1\times 1\times C$, converting them into the heatmaps of 50 intermediate virtual points; these heatmaps are in turn converted using a softargmax layer to $2C$ x-y pairs which are finally fed to a linear regressor to estimate manually annotated landmarks.
The parameters of the intermediate points and linear regressor are learned using a certain number of manual annotations, but the signal is not back-propagated further so the embeddings remain fully unsupervised.

In detail, after pretraining both the \textit{SmallNet} and \textit{Hourglass} networks on the CelebA dataset in a unsupervised manner, we freeze its parameters and only learn the regressors for MAFL~\cite{Zhang2016}.  We then follow the same methodology for the 68-landmark 300-W dataset~\cite{sagonas2013}, with 3148 training and 689 testing images.  We also evaluate on the challenging AFLW~\cite{koestinger2011} dataset, under the 5 landmark setting.  Two slightly different evaluation splits for have been used in prior work: one is the train/test partition of AFLW used in the works of \cite{thewlis17unsupervised}, \cite{thewlis17Bunsupervised} which used the existing crops from MTFL~\cite{zhang2014facial} and provides 2,995 faces for testing and 10,122 AFLW faces for training (we refer to this split as AFLW$_M$).  The second is a set of re-cropped faces released by \cite{zhang2018unsupervised}, which comprises 2991 test faces with 10,122 train faces (we refer to this split as AFLW$_R$).  For both AFLW partitions, and similarly to~\cite{thewlis17Bunsupervised}, after training for on CelebA we continue with unsupervised pretraining on 10,122 training images from AFLW for 50 epochs (we provide an ablation study to assess the effect of this choice in~\cref{subsec:ablations}).  We report the errors in percentage of inter-ocular distance in~\cref{t:landmark} and compare our results to state-of-the-art supervised and unsupervised methods, following the protocol and data selection used in~\cite{thewlis17Bunsupervised} to allow for a direct comparison.

We first see that the proposed DVE method outperforms the prior work that either learns sparse landmarks~\cite{thewlis17unsupervised} or 3D dense feature descriptors~\cite{thewlis17Bunsupervised}, which is consistent with the results in \cref{t:facematching}.  Encouragingly, we also see that our method is competitive with the state-of-the-art unsupervised learning techniques across the different benchmarks, indicating that our unsupervised formulation can learn useful information for this task.

\subsection{Ablations}\label{subsec:ablations}

In addition to the study evaluating DVE presented in table~\ref{t:facematching}, we conduct two additional experiments to investigate: (i) The sensitivity of the landmark regressor to a reduction in training annotations; (ii) the influence of additional unsupervised pretraining on a target dataset.\\

\begin{figure}[t]
\centering
\includegraphics[width=.5\textwidth]{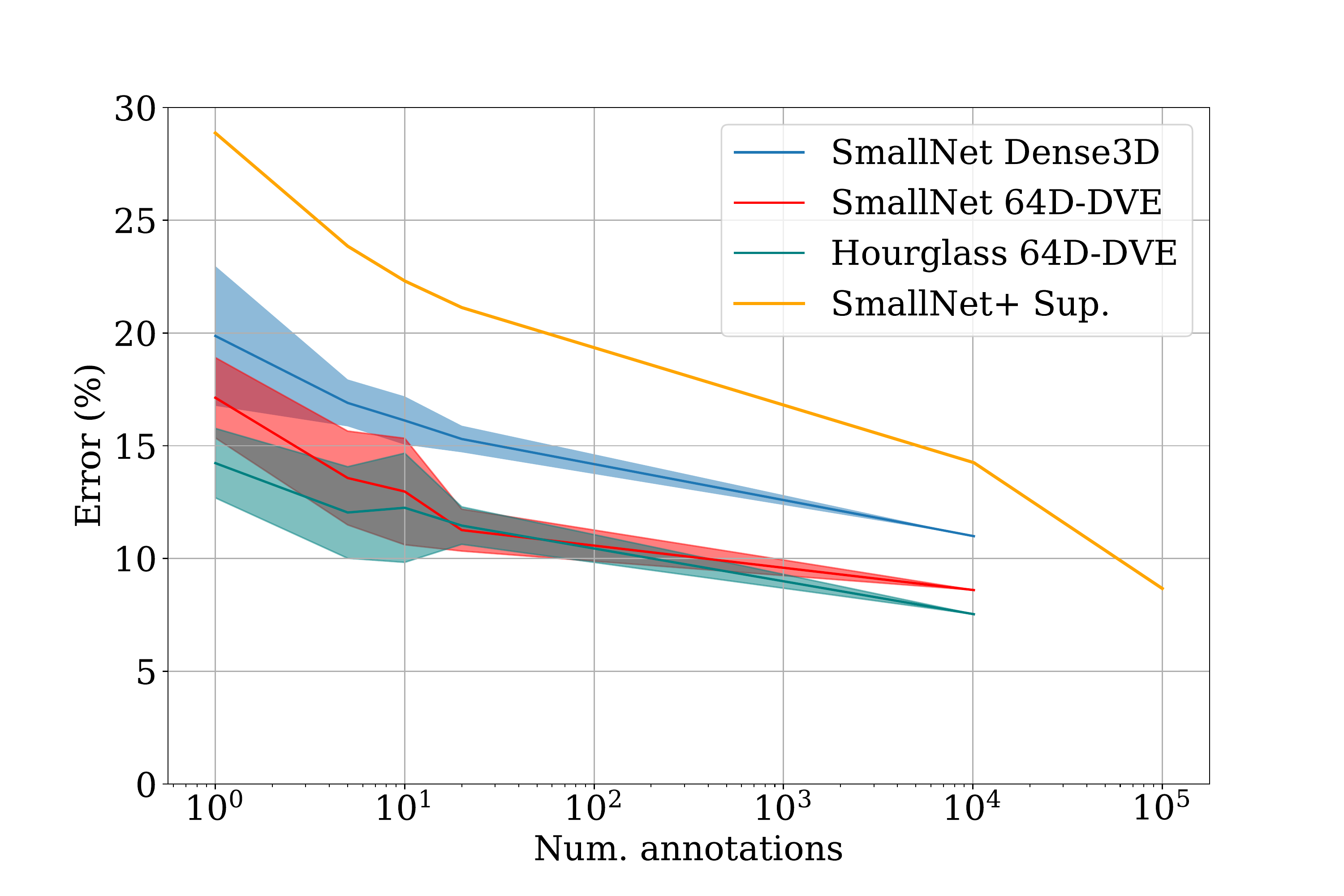}
\caption{The effect of varying the number of annotated images used for different methods on AFLW$_M$, incorporating the Supervised CNN baseline from~\cite{thewlis17unsupervised} (suppl. material). } 
\label{f:ntrainchart}
\vspace{-0.1cm}
\end{figure}

\noindent \textbf{Limited annotation:} We evaluate how many image annotations our method requires to learn landmark localisation in the AFLW dataset, comparing to Dense3D~\cite{thewlis17Bunsupervised} (which shares the SmallNet backbone architecture).
To do so, we vary the number of training images across the following range: 1, 5, 10, 20 and up to the whole training set (10,122 in total) and report the errors for each setting in~\cref{f:ntrainchart}.   For reference, we also include the supervised CNN baseline from~\cite{thewlis17unsupervised} (suppl. material), which consists of a slightly modified SmallNet (denoted SmallNet+ in fig.~\ref{f:ntrainchart}) to make it better suited for landmark regression.  Where available, we report the mean and std. deviation over three randomly seeded runs.  Further details of this experiment and the SmallNet+ architecture are provided in the suppl. material.  While there is considerable variance for very small numbers of annotations, the results indicate that DVE can produce effective landmark detectors with few manual annotations. \\

\noindent \textbf{Unsupervised finetuning:} Next we assess the influence of using unsupervised finetuning of the embeddings on a given target dataset, immediately prior to learning to regress landmarks.  To do so, we report the performance of several models with and without finetuning on both the AFLW$_M$ and 300W benchmarks in table~\ref{t:ftablation}.  We see that for AFLW$_M$, this approach (which can be achieved ``for free'' i.e. without collecting additional annotations) brings a boost in performance.  However, it is less effective for 300W, particularly at higher dimensions, having no influence on the performance of the stronger hourglass model.

\begin{table}[t]
\setlength\tabcolsep{6.2pt}
\centering
\begin{tabular}{cccc}
\toprule
Backbone & Embed. dim & AFLW$_M$ & 300W \\
\midrule
SmallNet & 3  & 11.82 / 11.12 & 7.66 / 7.20 \\
SmallNet & 16 & 10.22 / 9.15 & 6.29 / 5.90  \\
SmallNet & 32 & 9.80 / 9.17 & 6.13 / 5.75  \\
SmallNet & 64 & 9.28 / 8.60 & 5.75 / 5.58  \\
Hourglass & 64 & 8.15 /7.53 & 4.65 / 4.65  \\
\bottomrule
\end{tabular}
\medskip
\caption{The effect of unsupervised finetuning on landmark regression performance (errors reported as percentage of inter-ocular distance). Each table entry describes performance without/with finetuning.  All methods use DVE.}
\vspace{-0.4cm}
\label{t:ftablation}
\end{table}

\begin{figure*}[t!]
\centering
\includegraphics[width=.98\textwidth,keepaspectratio,clip,trim=0 7.3em 0 0]{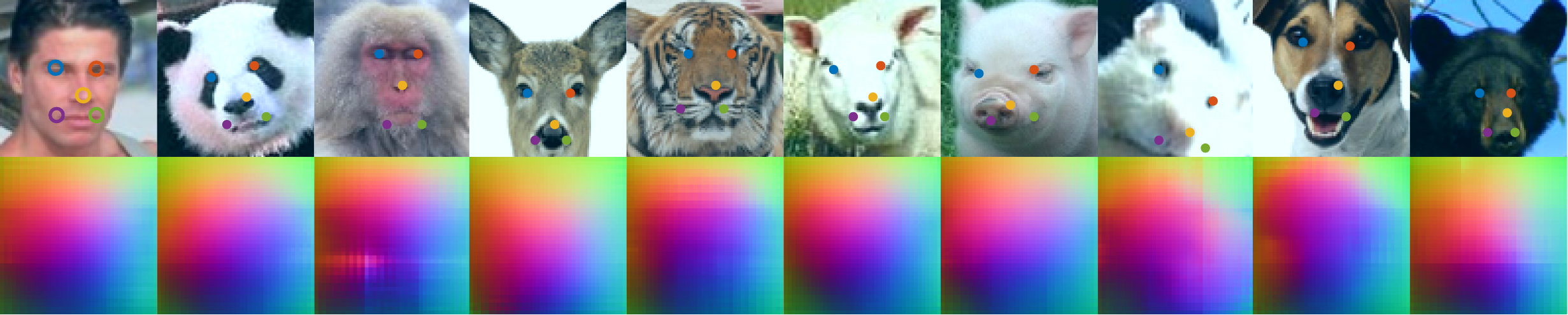}\vspace{-2pt}
\includegraphics[width=.98\textwidth,keepaspectratio,clip,trim=0 7.2em 0 0]{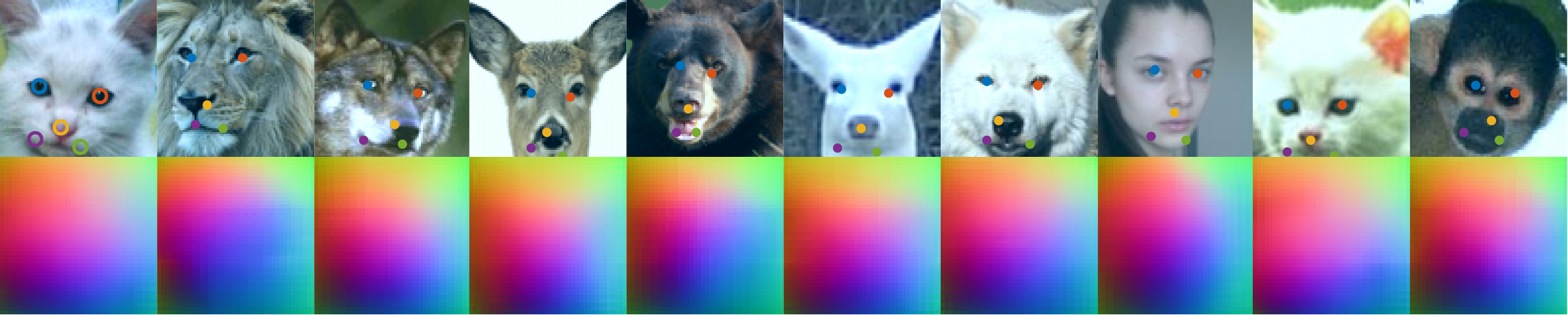}
\caption{Top: Five landmarks are manually annotated in the top-left image (human) and matched using our unsupervised embedding to a number of animals.
Bottom: same process, but using a cat image (bottom left) as query.}\label{f:animalfaces}
\vspace{-1em}
\end{figure*}

\subsection{Animal faces}\label{s:cross}
To investigate the generalisation capabilities of our method, we consider learning landmarks in an unsupervised manner not just for humans, but for animal faces.
To do this, we simply extend the set $\mathcal{X}$ of example image to contain images of animals as well.

In more detail, we consider the Animal Faces dataset~\cite{si2012learning} with images of 20 animal classes and about 100 images per class.
We exclude birds and elephants since these images have a significantly different appearance on average (birds profile, elephants include whole body).
We then add additional 8609 additional cat faces from~\cite{zhang2008cat}, 3506 cat and dog faces from~\cite{parkhi12a}, and 160k human faces from CelebA (but keep roughly the same distribution of animal classes per batch as the original dataset).
We train SmallNet descriptors using DVE on this data.  Here we also found it necessary to use the grouped attention mechanism (\cref{s:group}) which relaxes DVE to project embeddings on a set of auxiliary images rather than just one.
In order to do so, we include 16 pairs of images $(\bx,\bx')$ in each batch and we randomly choose a set of 5 auxiliary images for each pair from a separate pool of 16 images.
Note that these images have also undergone synthetic warps.  Results matching human and cat landmarks to other animals are shown in~\cref{f:animalfaces}.  DVE achieves localisation of semantically-analogous parts across species, with excellent results particularly for the eyes and general facial region.

\subsection{Roboarm}\label{s:robo}

\begin{figure}[h]
\centering
\includegraphics[width=\columnwidth,keepaspectratio,trim=0.4cm 0.2cm 0.4cm 0.12cm,clip]{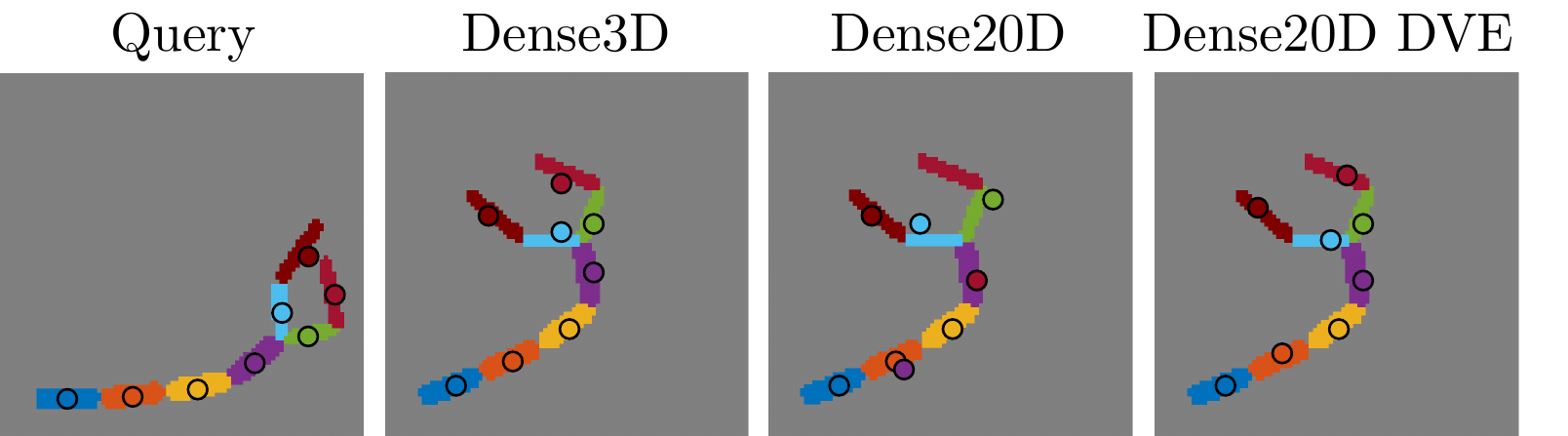}
\caption{An example of descriptor matching on a pair from the roboarm dataset, using blob centres in the first image to locate them in a second image.
We show 3D/20D descriptors (columns 2/3) learned with the loss from \cite{thewlis17Bunsupervised}.
The high error of the 20D case is corrected by DVE (last column).}\label{f:descmatchingroboarm}
\vspace{-0.25cm}
\end{figure}

\begin{table}[h]
\centering
\setlength{\tabcolsep}{0.5em}
	\begin{tabular}{cccc}
		\toprule
		Dimensionality    & \cite{thewlis17Bunsupervised} & + DVE   & - transformations \\
		\midrule
		3                 & 1.42                          & 1.41                & 1.69              \\
		20                & 10.34                         & 1.25                & 1.42              \\
		\bottomrule
	\end{tabular}
    \medskip
	\caption{Results on Roboarm, including an experiment ignoring optical flow (right).}
	\label{t:roboarmresults}
\end{table}

Lastly, we experimented on the animated robotic arm dataset (\cref{f:descmatchingroboarm}) introduced in~\cite{thewlis17Bunsupervised} to demonstrate the applicability of the approach to diverse data.
This dataset contains around 24k images of resolution $90\times 90$ with ground truth optical flow between frames for training.
We use the same matching evaluation of~\cref{s:human} using the centre of the robot's segments as keypoints for assessing correspondences.
We compare models using 3D and 20D embeddings using the formulation of~\cite{thewlis17Bunsupervised} with and without DVE, and finally removing transformation equivariance from the latter (by setting $g=1$ in~\cref{e:probmatch2}).

In this case there are no intra-class variations, but the high-degree of articulation makes matching non-trivial.

Without DVE, 20D descriptors are poor (10.34 error) whereas 3D are able to generalise (1.42). With DVE, however, the 20D descriptors (at 1.25 error) outperform the 3D ones (1.41).
Interestingly, DVE is effective enough that even removing transformations altogether (by learning from pairs of identical images using $g=1$) still results in good performance (1.42) -- this is possible because matches must hop through the auxiliary image set $\bx_\alpha$ which contains different frames.

\section{Conclusions}\label{s:conc}

We presented a new method that can learn landmark points in \emph{an unsupervised way}.
We formulated this problem in terms of finding correspondences between objects from the same or similar categories.
Our method bridges the gap between two seemingly independent concepts: landmarks and local image descriptors.
We showed that relatively high dimensional embeddings can be used to simultaneously match and align points by capturing instance-specific similarities as well as more abstract correspondences.
We also applied this method to predict facial landmarks in standard computer vision benchmarks as well as to find  correspondences across different animal species.
\paragraph{Acknowledgements.} We thank Almut Sophia Koepke for helpful discussions.  We are grateful to ERC StG IDIU-638009, EP/R03298X/1 and AWS Machine Learning Research Awards (MLRA) for support.
{\clearpage\bibliographystyle{ieee_fullname}\bibliography{refs}}

\clearpage
\onecolumn
\section*{Appendix}
\appendix

\begin{figure}[t]
	\centering
	\includegraphics[width=.95\textwidth,keepaspectratio]{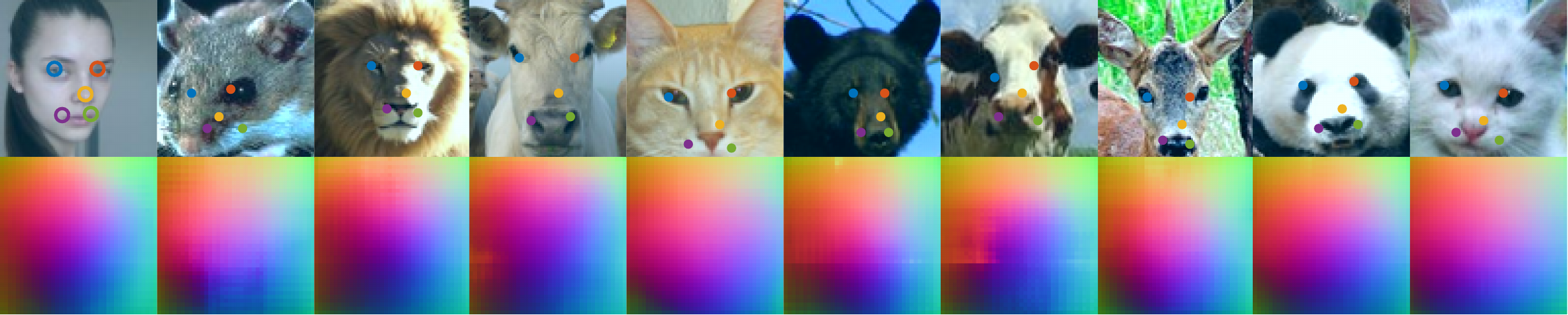}\vspace{-1pt}
	\includegraphics[width=.95\textwidth,keepaspectratio]{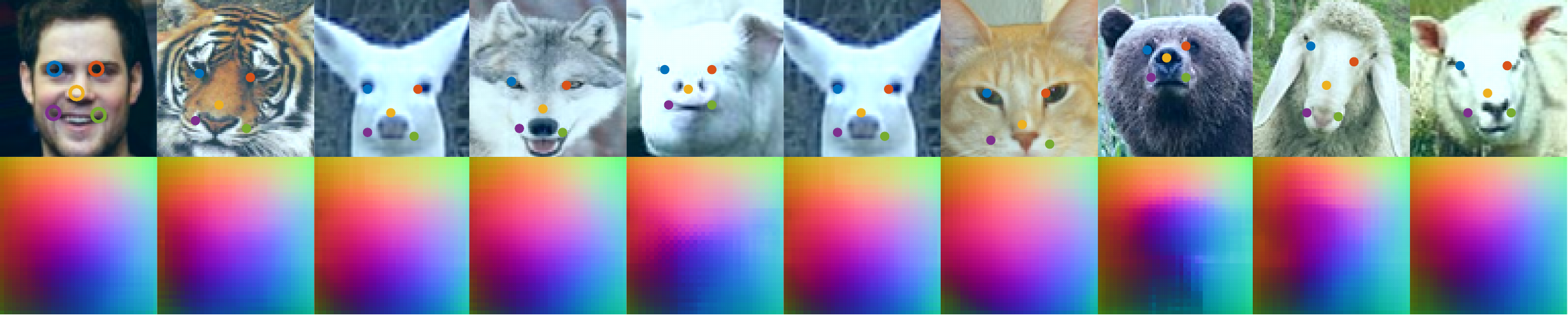}\vspace{-1em}
	\includegraphics[width=.95\textwidth,keepaspectratio]{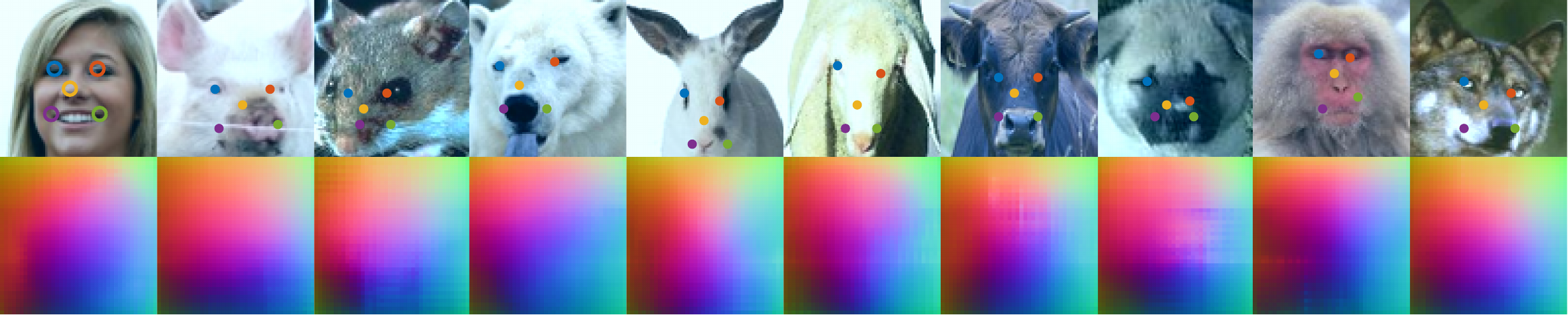}
	
	\caption{Additional images querying manual annotations on a human and finding the matching descriptors on animal faces, images are selected randomly from the validation set and include some more severe failure cases (eg mouse mouth row 1 col 2, dog eye row 5 col 8).}\label{f:animalfacesmore}
\end{figure}

\section{Animal Faces}

Here we present additional qualitative results on the animal faces dataset in~\cref{f:animalfacesmore}. The leftmost column shows the human faces with their annotated landmarks (drawn as coloured circles) which are matched to a set of queried animal faces in the remaining columns. The correspondent matches are depicted in the same colour with the manual landmarks. We observe that our method achieves to find many semantically meaningful matches in spite of wide variation in the appearance across different species.

\section{Roboarm details}

We showed results for an experiment showing that the use of optical flow (ground truth flow in this dataset) may not be essential. In this setup we set $\bx=\bx'$ when training, meaning the same image is used rather than two consecutive frames, and the transformation is the identity $g=1$. However we still explicitly ignore the background region (which is otherwise achieved by ignoring areas of zero flow). Our hope is that DVE, which has the effect of searching for a matching descriptor in a third image $\bx_\alpha$, will be able to stand in for explicit matches given by $g$. The results appear to confirm this. Surprisingly we can even obtain results matching~\cite{thewlis17Bunsupervised} without flow information, albeit using a higher dimensionality 20D descriptor. However the highest performance is still obtained by using the flow in addition to DVE, therefore we use flow (from synthetic warps) in experiments on faces.

\section{Limited Annotation experiments}

The numbers corresponding to the figure shown in section 4 of the paper portraying the effect of varying the number of annotated images are given in \cref{t:limited-annos}.  To allow the experiment to be reproduced, the list of the randomly sampled annotations is provided on the project page \url{http://www.robots.ox.ac.uk/~vgg/research/DVE}.

\begin{table*}[t]
\centering
\setlength\tabcolsep{0.5em}
\begin{tabular}{lcccc}
\toprule
Num.\ images & Dense 3D~\cite{thewlis17Bunsupervised}  &  SmallNet 64D DVE & Hourglass 64D DVE & Smallnet+ Sup. \\
\midrule
1                   & $19.87 \pm 3.10$ & $17.13 \pm 1.78$ & $14.23 \pm 1.54$ & 28.87 \\
5                   & $16.90 \pm 1.04$ & $13.57 \pm 2.08$ & $12.04 \pm 2.03$ & 32.85 \\
10                  & $16.12 \pm 1.07$ & $12.97 \pm 2.36$ & $12.25 \pm 2.42$ & 22.31\\
20                  & $15.30 \pm 0.59$ & $11.26 \pm 0.93$ & $11.46 \pm 0.83$ & 21.13 \\
AFLW$_M$ (10,122)   & 10.99          & 8.80           & 7.53           & 14.25\\
CelebA + AFLW$_M$ ($>$ 100k) & -     & -              & -               & 8.67 \\
\bottomrule
\end{tabular}
\medskip
\caption{Error (\% inter-ocular distance) Varying the number of images used for training (AFLW$_M$). The errors are reported in the form (mean $\pm$ std.), where the statistics are computed three randomly seeded samples of annotations. The general indication is that most of the information has been encoded in the unsupervised stage. }\label{t:limited-annos}
\end{table*}

\section{Architecture details}

The \textit{SmallNet} architecture, which was used in in~\cite{thewlis17unsupervised,thewlis17Bunsupervised} consists of a set of seven convolutional layers with 20, 48, 64, 80, 256 before $C$ filters are used to produce a $C$-dimensional embedding. The second, third and fourth layers are dilated by 2, 4 and 2 resp.  Every convolutional layer except the last is followed by batch norm and a {ReLU}. Following the first convolutional layer, features are downsampled via a $2\times 2$ max pool layer using a stride of 2.  Consequently, for an input size of $H\times W\times 3$, the size of the output $\frac{H}{2}\times \frac{W}{2}\times C$.

The \textit{SmallNet+} architecture, introduced in \cite{thewlis17unsupervised}, is a slightly modified version of SmallNet, which further includes pooling layers with a stride of 2 after each of the first three convolutional layers, and operates on an input of size $64 \times 64$.

The \textit{Houglass} architecture was introduced (in its ``stacked'' formulation) by \cite{newell2016stacked}.  We use this network with a single stack, operating on inputs of size $96 \times 96$ and using preactivation residuals.  

The code implementing the architectures used in this work can be found via the project page: \url{http://www.robots.ox.ac.uk/~vgg/research/DVE}.

\subsection{Preprocessing details}

For all datasets, the inputs to \textit{SmallNet} are then resized to $100\times100$ pixels then centre-cropped to $70\times70$ pixels (this cropping is done after warping during training).  The inputs to \textit{Hourglass} are resized to $136\times136$ pixels then centre-cropped to $96\times96$ pixels.  For the particular case of the CelebA face crops, which contain a good deal of surrounding context with varied backgrounds, faces are additionally preprocessed by removing the top $30$ pixels and bottom $10$ vertically from the $218\times178$ image before resizing. For 300-W we make the ground truth bounding box square (setting the height to equal the width) and then add more context on each side such that the original width occupies the central 52\% of the resulting image.

\end{document}